# Large Scale Variational Bayesian Inference for Structured Scale Mixture Models

**Young Jun Ko**                                                                                   YOUNGJUN.KO@EPFL.CH
**Matthias Seeger**                                                                              MATTHIAS.SEEGER@EPFL.CH
School of Computer and Communication Sciences, Ecole Polytechnique Fédérale de Lausanne, Switzerland

## Abstract

Natural image statistics exhibit hierarchical dependencies across multiple scales. Representing such prior knowledge in non-factorial latent tree models can boost performance of image denoising, inpainting, deconvolution or reconstruction substantially, beyond standard factorial "sparse" methodology. We derive a large scale approximate Bayesian inference algorithm for linear models with non-factorial (latent tree-structured) scale mixture priors. Experimental results on a range of denoising and inpainting problems demonstrate substantially improved performance compared to MAP estimation or to inference with factorial priors.

## 1. Introduction

Leaps in performance have been realized for low-level computer vision problems, such as denoising, inpainting, deconvolution (deblurring), image coding, undersampled reconstruction or acquisition optimization, by adopting super-Gaussian ("sparse") image priors. While most such methods employ simple factorial priors on single coefficients or groups, further substantial gains can be obtained by modelling higher-order dependencies via structured non-factorial prior distributions (Portilla et al., 2003; Wipf & Nagarajan, 2008; Cevher et al., 2010). For example, representing the dependencies among multi-scale wavelet coefficients by a (latent) tree structure can boost accuracy for image compression and reconstruction (Crouse et al., 1998; Papandreou et al., 2008; He et al., 2010). However, previous approaches employing such non-factorial priors either run much slower than standard factorial methodology, or sacrifice performance by adopting suboptimal MAP estimation or naive mean field factorization assumptions, ignoring posterior covariance or uncertainty altogether in problems which are highly underdetermined.

In this paper, we derive a large scale approximate Bayesian inference algorithm for generalized linear models with non-factorial (latent tree-structured) scale mixture priors. Our contributions are as follows:

- An image model with scale mixture prior on multi-scale wavelet coefficients, based on a latent discrete tree distribution. Mixture potentials can have arbitrary super-Gaussian components.

- A large scale double loop algorithm for Bayesian inference in these hybrid models. We do not require factorization assumptions between image pixels or wavelet coefficients. Our method is based on standard scalable technology (preconditioned conjugate gradients, penalized least squares) and can operate at the same scales as MAP estimation.

- An extension to incorporate non-log-concave potentials, such as Student's t, without sacrificing robustness of the optimization.

- Automatic Bayesian learning of a substantial number of hyperparameters from raw data. Folded into the variational optimization, this process does not require much overhead. It improves performance very significantly.

- An extensive evaluation on many inpainting and denoising datasets, comparing variational inference and MAP estimation for factorial and non-factorial priors featuring different sparsity potentials.

Our findings suggest (a) that predicting the variational *posterior mean* strongly and consistently outperforms the popular *posterior mode* (MAP estimation), (b) that non-factorial priors tend to boost performance compared to factorial ones, and (c) that Bayesian hyperparameter learning improves posterior mean prediction substantially compared to a default initialization.

The structure of the paper is as follows. We describe and motivate our image model in Section 2, and develop our large scale Bayesian inference and learning algorithm in Section 3. We present experimental results on image denoising and inpainting in Section 4, and close with conclusions.

### 1.1. Related Work

A range of prior work has employed latent tree-structured priors to represent dependencies between wavelet coefficients. Crouse et al. (1998) use Gaussian mixture potentials as well as a hidden Markov tree on the discrete mixture indicators, estimating latent signal and mixture parameters by expectation maximization. Portilla et al. (2003) employ continuous scale mixtures, based on a latent Gaussian tree mapped through coordinate-wise nonlinearities. They estimate parameters by nonlinear optimization. Papandreou et al. (2008) employ a hidden Markov tree and Gaussian mixture potentials on an overcomplete wavelet representation, estimating signal and parameters by Viterbi training. None of these employ Bayesian inference over the image or non-Gaussian potentials. He et al. (2010) use spike-and-slab prior potentials with a hidden Markov tree over the indicators. They perform standard naive mean field variational inference, employing a posterior distribution which does not represent any dependencies between coefficients. These assumptions lead to simple update equations, which they iterate in parallel. Their algorithm does not seem to reduce to standard scalable optimization primitives. Moreover, their method does not extend beyond spike-and-slab to other sparsity potentials.

Our framework can be seen as extension of the scalable double loop algorithm for variational inference in super-Gaussian models proposed in (Seeger & Nickisch, 2011). However, they did not consider hybrid models (featuring discrete and continuous variables), latent tree non-factorial prior distributions, or Bayesian learning of hyperparameters, as we do here.

## 2. Structured Image Model

In problems such as image denoising, inpainting or deconvolution (debluring), we seek to reconstruct a latent image $\boldsymbol{u} \in \mathbb{R}^n$ (an $n_y \times n_x$ bitmap, where $n = n_y n_x$) from noisy linear measurements $\boldsymbol{y} \in \mathbb{R}^m$:

$$\boldsymbol{y} = \boldsymbol{X}\boldsymbol{u} + \boldsymbol{\varepsilon}, \quad \boldsymbol{\varepsilon} \sim N(\boldsymbol{0}, \sigma^2 \boldsymbol{I}),$$

or $P(\boldsymbol{y}|\boldsymbol{u}) = N(\boldsymbol{y}|\boldsymbol{X}\boldsymbol{u}, \sigma^2 \boldsymbol{I})$, where $\sigma^2 > 0$ is the noise variance. For example, $\boldsymbol{X} = \boldsymbol{I}$ for denoising, $\boldsymbol{X} = \boldsymbol{I}_{J,\cdot}$ for inpainting ($J$ the index of observed pixel positions), or $\boldsymbol{X}\boldsymbol{u} = \boldsymbol{k} * \boldsymbol{u}$ for deconvolution ($\boldsymbol{k}$ the blur kernel). With $m \leq n$ (more unknowns than measurements), noise and/or blur, only additional statistical information in form of a *image prior distribution* $P(\boldsymbol{u})$ renders image reconstruction a well-posed problem. Given $P(\boldsymbol{u})$, we can infer the image from the *posterior distribution*

$$P(\boldsymbol{u}|\boldsymbol{y}) = \frac{P(\boldsymbol{y}|\boldsymbol{u})P(\boldsymbol{u})}{P(\boldsymbol{y})}, \quad P(\boldsymbol{y}) = \int P(\boldsymbol{y}|\boldsymbol{u})P(\boldsymbol{u})\,d\boldsymbol{u}. \tag{1}$$

For example, image statistics tend to be leptokurtic (super-Gaussian, or "sparse"), and even simple factorial image priors $P(\boldsymbol{u})$ respecting these properties can lead to dramatic improvements over classical methodology (least squares, Wiener filtering). Beyond marginals, image statistics exhibit complex dependencies, and capturing these in non-factorial priors can lead to further leaps in performance (Portilla et al., 2003). However, with probabilistic inference becoming much more difficult and expensive, such extended models enjoy little popularity so far compared to simpler factorial alternatives. In contrast, the methodology[1] developed in this paper scales up in the same way as MAP estimation and variational inference for factorial priors.

Consider an orthonormal discrete wavelet transform $\boldsymbol{B}$, and denote corresponding wavelet coefficients by $\boldsymbol{s} = \boldsymbol{B}\boldsymbol{u}$. A factorial image prior has the form $P(\boldsymbol{u}) \propto \prod_{j=1}^{n} t_j(s_j)$, where common super-Gaussian potentials include *Laplacian*

$$t_j(s_j) \propto \tau_j e^{-\tau_j |s_j|}, \quad \tau_j > 0, \tag{2}$$

or *Student's t* potentials

$$t_j(s_j) \propto \tau_j^{1/2} \left(1 + (\tau_j/\nu)s_j^2\right)^{-(\nu+1)/2}, \quad \tau_j, \nu > 0. \tag{3}$$

In the sequel, we assume that $\int t_j(s_j)\,ds_j = 1$. Each coefficient $s_j$ belongs to a scale level of analysis $l(j) \in \{1, \ldots, L\}$, $l = 1$ the coarsest, $l = L$ the finest scale (assume that $n_y$, $n_x$ are multiples of $2^L$). Even though the $s_j$ are approximately uncorrelated for natural images, it is well known that there are substantial causal dependencies between coefficients in neighbouring levels (Portilla et al., 2003). These are typically mod-elled by a quad-tree[2] $\mathcal{T}$, linking a coefficient at level $l < L$ to four children at level $l + 1$. For example, energy localized at a fine scale coefficient percolates its way up the tree through coarser scales. In order to capture this signature in a non-factorial prior, we use binary variables $\delta_j \in \{0, 1\}$, one for each $s_j$, as well as mixture potentials $t_j(s_j; \delta_j) = t_{0j}(s_j)^{1-\delta_j} t_{1j}(s_j)^{\delta_j}$, where $t_{0j}(s_j)$ enforces $s_j \approx 0$ more strongly than $t_{1j}(s_j)$. Each coefficient $s_j$ can be in low ($\delta_j = 0$) or high ($\delta_j = 1$) state, with $|s_j|$ penalized accordingly. Moreover, we use a prior $P(\boldsymbol{\delta})$ of directed graphical quad-tree structure $\mathcal{T}$, which encourages the inheritance of high/low states from a parent node to its children: $P(\boldsymbol{\delta}) = \prod_j P(\delta_j | \delta_{\pi(j)})$, where $\pi(j)$ is the parent node of $j$ in the quad-tree. All nodes of a level share a common conditional probability table, so that $P(\boldsymbol{\delta})$ is parameterized by $2L$ hyperparameters $\theta_{r,l} \equiv P(\delta_j = 1 | \delta_{\pi(j)} = r)$ for $l(j) = l$. The implied prior on the image

$$P(\boldsymbol{u}) = \sum_{\boldsymbol{\delta}} \prod_{j=1}^{n} \underbrace{t_{0j}(s_j)^{1-\delta_j} t_{1j}(s_j)^{\delta_j}}_{=:t_j(s_j;\delta_j)} P(\boldsymbol{\delta}), \quad \boldsymbol{s} = \boldsymbol{B}\boldsymbol{u}, \tag{4}$$

is a non-factorial scale mixture model, which faithfully represents the causal inheritance between wavelet coefficient sizes from coarse to fine levels. The normalization constant of $P(\boldsymbol{u})$ is one, since $\boldsymbol{B}$ is an orthonormal transform.

---

[1] Our model is somewhat simpler than that of (Portilla et al., 2003).

In Figure 1, we illustrate the effect a non-factorial prior (4) on results for an inpainting problem (75% pixels removed). The latent tree prior employs Student's t potentials $t_{1j}(s_j)$ for the high, Gaussian potentials $t_{0j}(s_j) = N(0, \xi_{0l(j)}^{-1})$ for the low state, where hyperparameters $\tau_{1l}, \xi_{0l}$ (two at each level) are learned automatically (results in third, $\delta_j$ marginals in fourth row). We also show results for a factorial prior with Student's t potentials for comparison (second row), whose hyperparameters $\tau_l$ are learned in the same way. The non-factorial prior leads to a more faithful reconstruction of wavelet coefficient at coarser scales, which motivates superior inpainting results in Section 4.

The outcome of our procedure is an approximate posterior distribution $Q(\boldsymbol{u}|\boldsymbol{y})Q(\boldsymbol{\delta}|\boldsymbol{y})$. The covariance of $Q(\boldsymbol{u}|\boldsymbol{y})$ can be used for decision making, e.g. Bayesian experimental design (Seeger & Nickisch, 2011), and our results indicate that it helps the hyperparameter learning. Moreover, we predict the (approximate) posterior mean $\mathrm{E}[\boldsymbol{u}|\boldsymbol{y}]$, not the posterior mode (MAP estimation). Our experiments demonstrate that the

---

[2] $\mathcal{T}$ is the computation tree for the fast Haar wavelet transform and describes dominating dependencies also for wavelets with larger support.

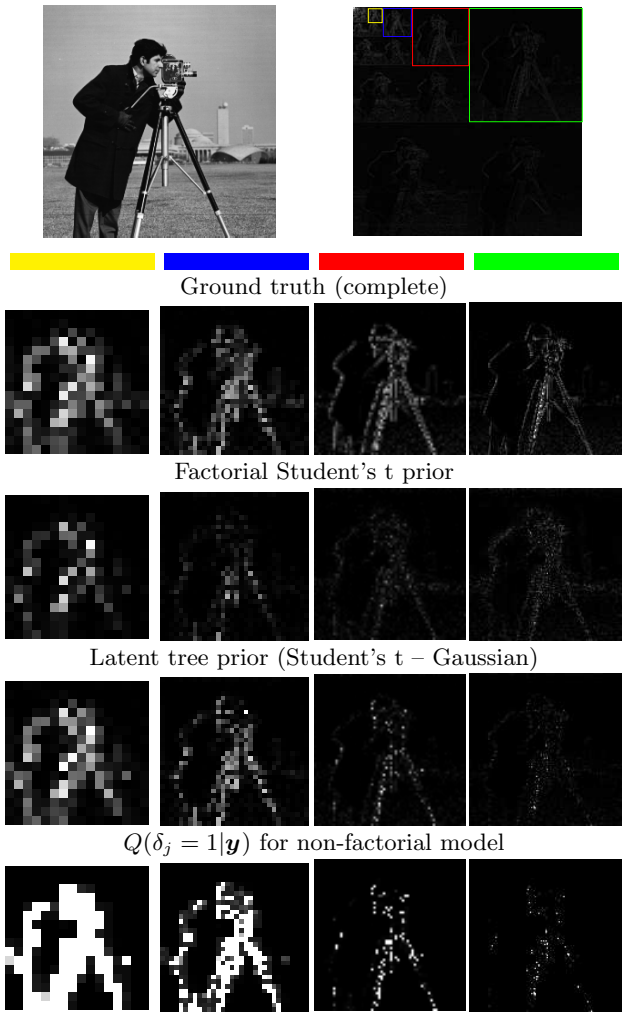

Figure 1. Shown are horizontal filterbands at levels $l = 5, \ldots, 8$ (left to right) of wavelet transform (top, right) of a natural image (top, left), both for complete image (first row) and inpainting results from 25% of pixels (second, third row). First row: While transform is sparse, large coefficients occur at all scales, localized clustering across levels is clearly visible. Second row: A factorial prior can lead to suppression of energy in coarser scales ($l = 6, 7$). Third row: A latent tree scale mixture prior represents wavelet statistics much better, energy at all levels is well reconstructed. Fourth row: Corresponding marginals $Q(\delta_j = 1|\boldsymbol{y})$ track large coefficients at all levels.

variational posterior mean leads to superior results in strongly ill-posed problems.

## 3. Large Scale Variational Inference

In this section, we derive a scalable algorithm for computing variational approximations to the posterior (1) for a non-factorial scale mixture prior (4). There are obviously strong dependencies between components of $\boldsymbol{u}$ or $\boldsymbol{s}$, and in contrast to previous work (He et al., 2010), we do not require any factorization assumptions between them. The high level idea behind our approach is *iterative decoupling*. We combine a standard variational bound to decouple $\boldsymbol{u}$ and $\boldsymbol{\delta}$ (allowing us to tackle inference over the latter by belief propagation) with the double loop framework of (Seeger & Nickisch, 2011), which decouples mean and covariance computations over $\boldsymbol{u}$. The latter provides a computational reduction to convex penalized least squares optimization and Gaussian sampling, which is crucial for scalability. As is shown below, minor simplifications result in inference or maximum a posteriori (MAP) estimation algorithms for non-factorial or factorial priors, all based on the same underlying code.

For the purpose of image priors, we restrict ourselves to even, *super-Gaussian* potentials $t_{rj}(s_j)$ (Palmer et al., 2006), which can be represented as

$$-2\log t(s) = \min_{\gamma \geq 0} s^2/\gamma + h(\gamma). \quad (5)$$

Intuitively, $t(s)$ can be tightly lower bounded by (unnormalized) Gaussians of any variance $\gamma$. Here, $h(\gamma)$ is convex if and only if $-2\log t(s)$ is convex (Seeger & Nickisch, 2011). Both Laplacian (2) and Student's t potentials (3) are super-Gaussian, while $-2\log t(s)$ is convex for Laplacian, but not for Student's t potentials. The criterion we will minimize is an upper bound on the negative log marginal likelihood $-2\log P(\boldsymbol{y})$ from (1). First, we introduce $\boldsymbol{\gamma}$ from (5), consisting of $\gamma_{0j}$ for $t_{0j}, h_{0j}$ and $\gamma_{1j}$ for $t_{1j}, h_{1j}$, and pull $\min_{\boldsymbol{\gamma}}$ out of the integral:

$$-2\log P(\boldsymbol{y}) = -2\log \int P(\boldsymbol{y}|\boldsymbol{u})P(\boldsymbol{u})\,d\boldsymbol{u}$$
$$\leq \min_{\boldsymbol{\gamma}} -2\log \sum_{\boldsymbol{\delta}} \int P(\boldsymbol{y}|\boldsymbol{u})e^{-\frac{1}{2}\boldsymbol{s}^T(\text{diag}\,\boldsymbol{\pi})\boldsymbol{s}}P(\boldsymbol{\delta})\,d\boldsymbol{u}$$
$$+ \underbrace{\sum_j (1-\delta_j)h_{0j}(\gamma_{0j}) + \delta_j h_{1j}(\gamma_{1j})}_{=:h(\boldsymbol{\gamma};\boldsymbol{\delta})},\quad \pi_j := \frac{1-\delta_j}{\gamma_{0j}} + \frac{\delta_j}{\gamma_{1j}},$$

where $\boldsymbol{s} = \boldsymbol{B}\boldsymbol{u}$. Second, we apply the variational mean field bound to the remaining log partition function:

$$\max_{Q(\boldsymbol{u}|\boldsymbol{y}),Q(\boldsymbol{\delta}|\boldsymbol{y})} E_Q\left[\log \frac{P(\boldsymbol{y}|\boldsymbol{u})e^{-\frac{1}{2}\boldsymbol{s}^T(\text{diag}\,\boldsymbol{\pi})\boldsymbol{s}}P(\boldsymbol{\delta})}{Q(\boldsymbol{u}|\boldsymbol{y})Q(\boldsymbol{\delta}|\boldsymbol{y})}\right],$$

using the factorization assumption $Q(\boldsymbol{u},\boldsymbol{\delta}|\boldsymbol{y}) = Q(\boldsymbol{u}|\boldsymbol{y})Q(\boldsymbol{\delta}|\boldsymbol{y})$. In the sequel, we denote $E_{Q(\boldsymbol{\delta}|\boldsymbol{y})}[\cdot]$ by $\langle \cdot \rangle$. The bound maximizer is $Q(\boldsymbol{u}|\boldsymbol{y}) = Z_Q(\langle\boldsymbol{\pi}\rangle)^{-1}P(\boldsymbol{y}|\boldsymbol{u})e^{-\frac{1}{2}\boldsymbol{s}^T(\text{diag}\langle\boldsymbol{\pi}\rangle)\boldsymbol{s}}$, where $Z_Q(\langle\boldsymbol{\pi}\rangle)$ is a Gaussian partition function. Plugging this in, we obtain the bound

$$\min_{Q(\boldsymbol{\delta}|\boldsymbol{y}),\boldsymbol{\gamma}} -2\log Z_Q(\langle\boldsymbol{\pi}\rangle) + h(\boldsymbol{\gamma};\langle\boldsymbol{\delta}\rangle) + 2\text{D}[Q(\boldsymbol{\delta}|\boldsymbol{y})\,\|\,P(\boldsymbol{\delta})],$$

using that $h(\boldsymbol{\gamma};\boldsymbol{\delta})$ is linear in $\boldsymbol{\delta}$. The relative entropy term is defined as $\text{D}[Q(\boldsymbol{\delta}|\boldsymbol{y})\,\|\,P(\boldsymbol{\delta})] = \langle \log Q(\boldsymbol{\delta}|\boldsymbol{y}) - \log P(\boldsymbol{\delta})\rangle$. Finally, and crucially for scalability, we use the variational transformation from (Seeger & Nickisch, 2011): $-2\log Z_Q$ is equal to

$$\min_{\boldsymbol{u}_*,\boldsymbol{z}} \underbrace{(\boldsymbol{z}+\boldsymbol{s}_*^2)^T\langle\boldsymbol{\pi}\rangle + \sigma^{-2}\|\boldsymbol{y}-\boldsymbol{X}\boldsymbol{u}_*\|^2}_{=:R_{\boldsymbol{z}}(\boldsymbol{u}_*,\langle\boldsymbol{\pi}\rangle)} - g^*(\boldsymbol{z}),$$

where $\boldsymbol{s}_* = \boldsymbol{B}\boldsymbol{u}_*$, and $\boldsymbol{z} \succ \boldsymbol{0}$. If $\boldsymbol{A}(\boldsymbol{\pi}) := \sigma^{-2}\boldsymbol{X}^T\boldsymbol{X} + \boldsymbol{B}^T(\text{diag}\,\boldsymbol{\pi})\boldsymbol{B}$ denotes the inverse covariance matrix of $Q(\boldsymbol{u}|\boldsymbol{y})$, then $\boldsymbol{\pi} \mapsto \log|\boldsymbol{A}(\boldsymbol{\pi})|$ is concave, and the $-2\log Z_Q$ representation is based on the corresponding Fenchel duality. Since the dependence of $\boldsymbol{A}$ on $\boldsymbol{\delta}$ is through $\boldsymbol{\pi}$, neither $\boldsymbol{z}$ nor $g^*(\boldsymbol{z})$ depends on $\boldsymbol{\delta}$. Plugging this in and pulling $\min_{\boldsymbol{u}_*,\boldsymbol{z}}$ outside, we obtain our final upper bound on $-2\log P(\boldsymbol{y})$:

$$\phi = R_{\boldsymbol{z}}(\boldsymbol{u}_*,\langle\boldsymbol{\pi}\rangle) + h(\boldsymbol{\gamma};\langle\boldsymbol{\delta}\rangle) + 2\text{D}[Q(\boldsymbol{\delta}|\boldsymbol{y})\,\|\,P(\boldsymbol{\delta})] - g^*(\boldsymbol{z}),$$

to be minimized over $Q(\boldsymbol{\delta}|\boldsymbol{y})$, $\boldsymbol{z}$, $\boldsymbol{u}_*$, $\boldsymbol{\gamma}$. Notice that $\langle\boldsymbol{\pi}(\boldsymbol{\delta})\rangle = \boldsymbol{\pi}(\langle\boldsymbol{\delta}\rangle)$ by linearity.

We adapt the scalable convergent double loop algorithm from (Seeger & Nickisch, 2011). During the inner loop, we fix $\boldsymbol{z}$ and minimize $\phi$ over $Q(\boldsymbol{\delta}|\boldsymbol{y})$ and $(\boldsymbol{u}_*,\boldsymbol{\gamma})$. In between, once per outer loop iteration, we update $\boldsymbol{z}$ to obtain a tangential fit to $\log|\boldsymbol{A}(\langle\boldsymbol{\pi}\rangle)|$: $\boldsymbol{z} \leftarrow \text{Var}_Q[\boldsymbol{s}|\boldsymbol{y}]$, where $Q(\boldsymbol{u}|\boldsymbol{y})$ is based on $\langle\boldsymbol{\pi}\rangle$ for the current $\boldsymbol{\gamma}$ and $\langle\boldsymbol{\delta}\rangle$. Computing these Gaussian variances is the most computationally intensive part of a variational inference method. We approximate $\boldsymbol{z}$ using the Perturb&MAP technique from (Papandreou & Yuille, 2010), at the cost of solving a small number of linear systems $\boldsymbol{A}(\langle\boldsymbol{\pi}\rangle)\boldsymbol{x}_k = \boldsymbol{r}_k$ by preconditioned conjugate gradients.

In the inner loop, we update $Q(\boldsymbol{\delta}|\boldsymbol{y})$ and $(\boldsymbol{u}_*,\boldsymbol{\gamma})$ alternatingly. For the latter, we can eliminate $\boldsymbol{\gamma}$ by reversing the super-Gaussian representation of $-\log t_{rj}(s_{*j})$, plugging in $p_j := (z_j + (s_{*j})^2)^{1/2}$ instead of $s_{*j}$. Dropping terms independent of $\boldsymbol{u}_*$, we need to solve

$$\min_{\boldsymbol{u}_*} \sigma^{-2}\|\boldsymbol{y}-\boldsymbol{X}\boldsymbol{u}_*\|^2 - 2\sum_j \log t_j\left(p_j;\langle\delta_j\rangle\right),$$
$$p_j = \sqrt{z_j + (s_{*j})^2},\quad \boldsymbol{s}_* = \boldsymbol{B}\boldsymbol{u}_*. \quad (6)$$

Here, we used that $\log t_j(s_j;\delta_j)$ is linear in $\delta_j$. This is a standard-form penalized least squares problem, which can be solved by any of a large number of recent algorithms developed for MAP estimation. In our experiments, we employ a nonlinear conjugate gradients algorithm from the `glm-ie` toolbox (see Section 4). Importantly, this inner loop problem is convex iff the $-\log t_{rj}(s_j)$ are convex, thus iff MAP estimation is convex for the same model (the precise relationship to MAP is detailed shortly). It is not convex if Student's t potentials (3) are used, and we will use additional bounding in order to retain inner loop convexity (Section 3.3). For the update of $Q(\boldsymbol{\delta}|\boldsymbol{y})$, note that

$$\phi \doteq \left\langle -2\sum_j \log t_j(p_j;\delta_j) \right\rangle + 2\mathrm{D}[Q(\boldsymbol{\delta}|\boldsymbol{y}) \,\|\, P(\boldsymbol{\delta})]$$
$$\doteq 2\left\langle \log \frac{Q(\boldsymbol{\delta}|\boldsymbol{y})}{P(\boldsymbol{\delta})\prod_j e^{\log t_j(p_j;\delta_j)}} \right\rangle,$$

where "$\doteq$" denotes equality up to an additive constant. We can read off the minimizer $Q(\boldsymbol{\delta}|\boldsymbol{y}) \propto P(\boldsymbol{\delta})\prod_j t_j(p_j;\delta_j)$. This is a distribution of the same quad-tree structure $\mathcal{T}$ as $P(\boldsymbol{\delta})$, differing from the latter in the single node potentials only. We can compute both $\langle\boldsymbol{\delta}\rangle$ and $\mathrm{D}[Q(\boldsymbol{\delta}|\boldsymbol{y}) \,\|\, P(\boldsymbol{\delta})]$ in $O(n)$, using Pearl's belief propagation algorithm.

This completes the description of our large scale inference algorithm. At convergence, $\boldsymbol{u}_*$ constitutes our posterior mean prediction $\mathrm{E}_Q[\boldsymbol{u}|\boldsymbol{y}]$. Apart from simple direct primitives, it reduces entirely to solving a moderate number of linear systems $\boldsymbol{A}(\boldsymbol{\pi})\boldsymbol{x} = \boldsymbol{r}$ for different $(\boldsymbol{\pi},\boldsymbol{r})$, which can be done very efficiently by state-of-the-art preconditioned conjugate gradients solvers. Our double loop structure implies that the most expensive updates of $\boldsymbol{z}$ have to be done least frequently.

### 3.1. MAP Estimation. Factorial Priors

We derived an algorithm for variational Bayesian inference with a structured non-factorial prior (4). Importantly, we can obtain scalable algorithms for MAP estimation or inference with factorial or non-factorial priors by making minor simplifying modification, otherwise *using exactly the same underlying code*. First, when using a factorial prior of the form $P(\boldsymbol{u}) = \prod_j t_j(s_j)$, we simply eliminate $t_{1j}$, set $\langle \delta_j \rangle = 0$ and eliminate $Q(\boldsymbol{\delta}|\boldsymbol{y})$ altogether. The inner loop now consists of a single penalized least squares problem (6). Second, MAP estimation is obtained by simply setting $\boldsymbol{z} = \boldsymbol{0}$, skipping variances computations and running a single outer loop iteration. MAP estimation for a non-factorial prior proceeds in an expectation-maximization fashion, alternating between penalized least squares (PLS) and belief propagation (Crouse et al., 1998).

### 3.2. Learning Prior Hyperparameters

In order to obtain best performance, it is necessary to endow an image prior $P(\boldsymbol{u})$ (whether factorial or not) with a substantial number of hyperparameters, which have to be adjusted to the problem at hand. For example, wavelet coefficients $s_j$ exhibit higher variance at coarser than at finer scales $l(j)$, and corresponding prior potentials $t_j(s_j)$ should take this into account, via $\tau_j$ in (2) or (3), or $\xi_j$ in Gaussians $N(0, \xi_j^{-1})$. In our experiments with non-factorial priors, we use $L = 8$ scale levels, giving rise to 16 hyperparameters (one for each level $l$ and low/high state $r$): too many to be reasonably set by non-Bayesian methods like cross-validation. In this section, we show how to learn hyperparameters in an automatic Bayesian way. Our method is folded into the variational inference process, which lets us optimize hyperparameters for each dataset at hand. We stress that Bayesian learning operates on the raw data (noisy, incomplete, blurred): clean underlying images are not required.

Bayesian learning works by maximizing the log marginal likelihood $\log P(\boldsymbol{y})$ w.r.t. hyperparameters. The obvious variational approximation is to maximize the lower bound (or, equivalently, minimize $\phi$) instead. Indeed, we can treat the hyperparameters (say, $\boldsymbol{\theta}$) as just another set of parameters to minimize $\phi$ over, thereby folding the learning into the inference approximation. Importantly, we can update $\boldsymbol{\theta}$ as part of our inner loop optimization, for fixed $\boldsymbol{z}$, without compromising overall convergence (to a stationary point). Recall that $\boldsymbol{z}$ comes from the Fenchel duality $\log |\boldsymbol{A}(\langle\boldsymbol{\pi}\rangle)| = \min_{\boldsymbol{z}} \boldsymbol{z}^T\langle\boldsymbol{\pi}\rangle - g^*(\boldsymbol{z})$. Since $\boldsymbol{A}$ depends on all other parameters[3] only through $\langle\boldsymbol{\pi}\rangle$, there is no direct dependence between $\boldsymbol{z}$ and $\boldsymbol{\theta}$. Suppose that all potentials $t_{rj}(s_j)$ for fixed $r \in \{0,1\}$ and $l = l(j)$ share a hyperparameter $\tau_{rl}$. Denote $q_{rj} := Q(\delta_j = r|\boldsymbol{y})$. We write "$j:l$" short for "$j:l(j) = l$". The relevant part of $\phi$ is

$$\phi \doteq -2 \sum_{j:l} q_{rj} \log t_{rj}(p_j), \quad p_j = \sqrt{z_j + (s_{*j})^2}.$$

We need to minimize $\phi$ w.r.t. $\tau_{rl}$, which can often be done analytically. For Laplacian potentials (2):

$$\phi \doteq 2 \sum_{j:l} q_{rj} \left( \tau_{rl} p_j - \log \tau_{rl} \right) \;\Rightarrow\; \tau_{rl} \leftarrow \frac{\sum_{j:l} q_{rj}}{\sum_{j:l} q_{rj} p_j}.$$

For Gaussian potentials $t_{rj}(s_j) \propto \xi_{rl}^{1/2} e^{-\frac{1}{2}\xi_{rl} s_j^2}$:

$$\phi \doteq \sum_{j:l} q_{rj} \left( \xi_{rl} p_j^2 - \log \xi_{rl} \right) \;\Rightarrow\; \xi_{rl} \leftarrow \frac{\sum_{j:l} q_{rj}}{\sum_{j:l} q_{rj} p_j^2}.$$

For Student's t potentials, we update hyperparameters

---

[3] An exception would be the noise variance $\sigma^2$. We can decouple $\log |\boldsymbol{A}(\langle\boldsymbol{\pi}\rangle)|$ w.r.t. $\sigma^2$ as well, but this is not done here. $\sigma^2$ is fixed in our experiments.

$\tau_{rl}$ only once per outer loop iteration, as detailed in Section 3.3.

Recall the parameterization of the tree prior $P(\boldsymbol{\delta})$ in terms of $\theta_{r,l}$ from Section 2. The relevant criterion part is $\phi \doteq \langle \log P(\boldsymbol{\delta}) \rangle$, which implies the updates

$$\theta_{r,l} \leftarrow \begin{cases} \frac{\sum_{j:1} q_{1j}}{\sum_{j:1} 1} & l = 1 \\ \frac{\sum_{j:l} q_{1rj}}{\sum_{j:l} (q_{1rj} + q_{0rj})} & l > 1 \end{cases}$$

where $q_{krj} := Q(\delta_j = k, \delta_{\pi(j)} = r | \boldsymbol{y})$, $k, r = 0, 1$, are double node marginals.

### 3.3. Student's T Potentials

When applied to models featuring Student's t potentials (3), the algorithm just detailed requires non-convex PLS problems (6) to be solved in the inner loop. Commonly used first order solvers can fail dramatically on non-convex problems. Since we adopt a double loop strategy anyway, it is simpler and far more robust to use additional bounding in order to obtain a convex inner loop problem. This idea has previously been described in (Seeger & Nickisch, 2011) and applied to Student's t potentials, but our applications here, as well as our hyperparameter learning method, are novel.

Recall the representation (5) of $t(s)$. For a Student's t potential, $h(\gamma)$ is not convex, but can be written as sum of a convex term $h_\cup(\gamma)$ and a concave term $h_\cap(\gamma)$ (Seeger & Nickisch, 2011, Appendix A.6). Moreover, the concave part can be represented by Fenchel duality: $h_\cap(\gamma) = \min_{e > 0} e\gamma - g_\cap^*(e)$. Overall, we end up with a further parameter vector $\boldsymbol{e} = [e_{rj}]$, which is updated alongside $\boldsymbol{z}$. Define $t_\cup(s; e)$ as

$$-2 \log t_\cup(s; e) = \min_{\gamma \geq 0} s^2/\gamma + h_\cup(\gamma) + e\gamma.$$

A minimum over jointly convex functions in $(s, \gamma)$, this is a convex function in $s$. We end up with a modified *convex* inner loop PLS problem of the form (6), where $-2 \log t_{rj}(p_j)$ is replaced by $-2 \log t_{\cup; rj}(p_j; e_{rj})$ for every Student's t potential $t_{rj}$. Here, $-2 \log t_\cup(s; e)$ is easily computed by a single case distinction. We have to replace $\log t_{rj}(p_j)$ by $\log t_{\cup; rj}(p_j; e_{rj})$ also in the update of $Q(\boldsymbol{\delta}|\boldsymbol{y})$.

The $\tau_{rl}$ hyperparameters for Student's t potentials are updated once per outer loop, alongside the $e_{rj}$. We use the same procedure as in Section 3.2, but applied to $-2 \log t_{rj}(p_j)$, not its convexification. The relevant criterion part is

$$\phi \doteq \sum_{j:l} q_{rj} \left( (\nu + 1) \log \left( 1 + (p_j^2/\nu) e^{\log \tau_{rl}} \right) - \log \tau_{rl} \right).$$

This is a smooth convex function in $\log \tau_{rl}$, which is easily minimized by a one-dimensional Newton solver. Once all Student's t hyperparameters have been updated, we refit the corresponding $e_{rj}$ and continue with another inner loop.

## 4. Experiments

We present experiments on a range of denoising and inpainting problems, comparing variational inference and MAP estimation for different models. Our results are averaged over 77 frequently used images (greyscale, $256 \times 256$), a dataset[4] from (Seeger & Nickisch, 2008). Our implementation is based on the `glm-ie` toolbox (www.mloss.org/software/view/269/). We compare 8 methods: MAP estimation (MAP) vs. variational inference (VB), factorial prior (fact) vs. latent tree scale mixture prior (tree), and Laplacian (Lap) vs. Student's t potentials (T). The *Lap-tree* model uses two Laplace potentials (2) $t_{0j}(s_j)$, $t_{1j}(s_j)$ with different hyperparameters $\tau_{0l}$, $\tau_{1l}$, a pair for each level. The *T-tree* model employs Gaussian $N(s_j|0, \xi_{0l}^{-1})$ for the low, Student's t potentials (3) for the high state, with a pair of hyperparameters $(\xi_{0l}, \tau_{1l})$ at each level. We use $2L$ hyperparameters in the *tree*, $L$ (namely, $\{\tau_l\}$) in the *fact* setups. The Student's t shape parameter $\nu$ is fixed to 2.1. For each run, we initialize hyperparameters $\boldsymbol{\theta}$ as in (Crouse et al., 1998), by maximizing the prior probability of the raw[5] data $\boldsymbol{y}$ (for the *tree* cases, this involves a few steps of expectation maximization), then optimize them by minimizing $\phi$. Hyperparameters were updated once per outer loop iteration.

VB runs use up to 15 outer loop (OL) iterations. Perturb&MAP estimation of $\boldsymbol{z}$, required for inpainting only, is run with 30 samples à 70 conjugate gradients (CG) iterations. We did 3 belief propagation and PLS calls per OL iteration for *tree* setups, PLS ran up to 150 iterations of nonlinear CG. Each iteration of CG requires two matrix-vector multiplications with $\boldsymbol{B}$ and $\boldsymbol{X}$. These choices have not been optimized for maximum efficiency.

### 4.1. Denoising

We add Gaussian random noise of variance $\sigma^2 = 0.01$ to each image (with pixel values $u_i \in [0, 1]$). All methods use the correct value of $\sigma^2$ in their likelihood. Notice that in this case, Gaussian variances $\boldsymbol{z}$ can be computed exactly at no cost. Namely,

---

[4] Thanks to H. Nickisch for providing the data. We added 2 further images.

[5] For inpainting, the missing pixels are set to mean($y_i$).

$A(\pi) = \sigma^{-2}I + B^T\Pi B$, where $B^T B = I$, so that

$$z = \mathrm{diag}^{-1}\left(BA(\pi)^{-1}B^T\right) = (\sigma^{-2}\mathbf{1} + \pi)^{-1}.$$

Therefore, Perturb&MAP, the dominating cost for VB in general, is not required. Results are shown in Table 1. For this application, differences between MAP and VB reconstruction are not significant. On the other hand, the non-factorial prior improves PSNR somewhat. Hyperparameter learning improves VB performance substantially, especially when Student's t potentials are used. In contrast, it does not help[6] (and can even hurt) MAP performance.

| Model | VB | | MAP | |
|---|---|---|---|---|
| | Init $\theta$ | Learned $\theta$ | Init $\theta$ | Learned $\theta$ |
| Lap-fact | $24.5 \pm 0.7$ | $\mathbf{24.7 \pm 0.9}$ | $24.5 \pm 1.4$ | $23.2 \pm 1.6$ |
| T-fact | $20.8 \pm 0.0$ | $23.3 \pm 0.5$ | $24.9 \pm 1.1$ | $24.7 \pm 1.6$ |
| Lap-tree | $24.3 \pm 0.7$ | $25.0 \pm 1.0$ | $\mathbf{25.1 \pm 1.3}$ | $25.0 \pm 1.6$ |
| T-tree | $21.1 \pm 0.1$ | $\mathbf{25.2 \pm 1.3}$ | $24.3 \pm 1.0$ | $24.3 \pm 1.6$ |

Table 1. Denoising experiments ($\sigma^2 = 0.01$). Shown is PSNR w.r.t. noise-free ground truth (mean and std.dev. over 77 images).

### 4.2. Inpainting

We remove 75% of pixels at random, using the same mask $J \subset \{1, \ldots, n\}$ for all images. The design matrix is $X = I_{J,\cdot}$, the noise variance was fixed to $\sigma^2 = 10^{-5}$. Results are shown in Table 2. As PSNR does not always correlate well with visual quality, we show a range of images in Figure 2, Figure 3, Figure 4.

| Model | VB | | MAP | |
|---|---|---|---|---|
| | Init $\theta$ | Learned $\theta$ | Init $\theta$ | Learned $\theta$ |
| Lap-fact | $23.0 \pm 2.1$ | $\mathbf{23.3 \pm 2.1}$ | $15.1 \pm 3.8$ | $20.0 \pm 2.3$ |
| T-fact | $21.0 \pm 1.7$ | $20.1 \pm 1.7$ | $\mathbf{20.6 \pm 2.2}$ | $20.0 \pm 2.2$ |
| Lap-tree | $22.2 \pm 2.0$ | $\mathbf{23.5 \pm 2.2}$ | $19.8 \pm 2.5$ | $19.7 \pm 2.6$ |
| T-tree | $21.2 \pm 1.8$ | $\mathbf{23.6 \pm 2.2}$ | $19.0 \pm 2.3$ | $19.8 \pm 2.5$ |

Table 2. Inpainting experiments (75% pixels removed). Shown is PSNR w.r.t. ground truth (mean and std.dev. over 77 images).

VB posterior mean predictions are clearly superior to MAP reconstruction, and VB with non-factorial latent tree prior performs best. While VB with a factorial Laplace prior (Lap-fact) shows similar PSNR values to VB-Lap-tree, the visual appearance of results with the latter is clearly superior (further results, provided in the supplemental material, support these findings). The additional runtime compared to MAP estimation, mainly due to the estimation of variances $z$, pays off for these problems.

---

[6] There is no justification for maximizing the posterior w.r.t. $\theta$. We include these results only for the fact that "alternating MAP" learning is frequently done in practice.

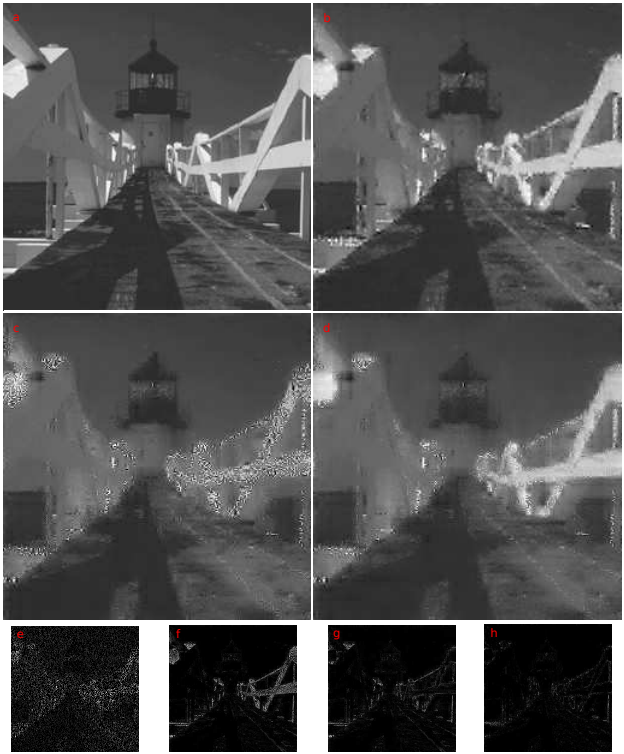

Figure 2. a: Truth. **b: VB-Lap-tree**. c: MAP-Lap-tree. d: VB-Lap-fact. e: $y$ (25%). f: $\Delta(c,a)$. g: $\Delta(d,a)$. **h:** $\Delta(b,a)$.

## 5. Conclusion

We presented a double loop algorithm for variational Bayesian inference in linear models with non-factorial scale mixture priors, based on a latent discrete tree distribution. Our method can operate at the same scales as MAP estimation, yet its (approximate) posterior mean prediction strongly and consistently outperforms the posterior mode across a range of inpainting problems. Both the selective smoothing by predictive variances and the coupling of wavelet coefficient across scales by way of the latent tree contribute to the removal of artefacts which plague results of MAP estimation, and of inference with factorial priors. Free hyperparameters are learned automatically by marginal likelihood maximization folded into the variational optimization.

In future work, we will try to adapt our large scale inference methodology to more complex hierarchical models, featuring non-Gaussian continuous latent variables (Portilla et al., 2003).

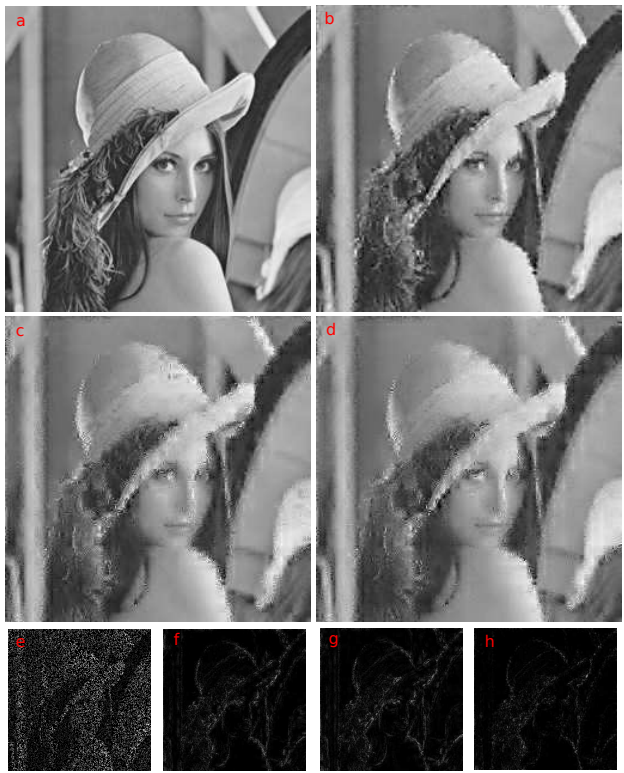

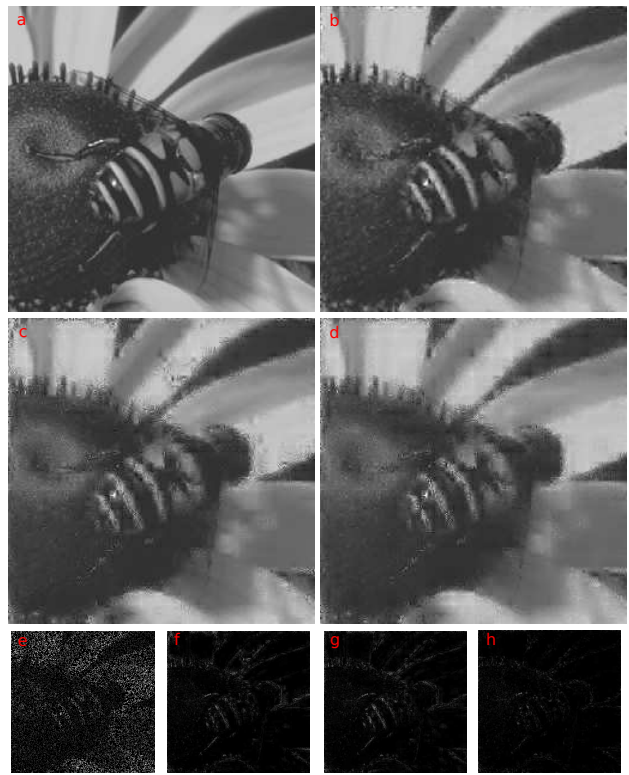

Figure 3. a: Truth. **b: VB-Lap-tree**. c: MAP-Lap-tree. d: VB-Lap-fact. e: $\boldsymbol{y}$ (25%). f: $\Delta(c,a)$. g: $\Delta(d,a)$. **h:** $\Delta(b,a)$.

Figure 4. a: Truth. **b: VB-Lap-tree**. c: MAP-Lap-tree. d: VB-Lap-fact. e: $\boldsymbol{y}$ (25%). f: $\Delta(c,a)$. g: $\Delta(d,a)$. **h:** $\Delta(b,a)$.

## Acknowledgments

Support through a DFG Sachbeihilfe SE 2008/1-1 (AOBJ 578593) and an ERC Starting Grant (277815 – SCALABIM) are gratefully acknowledged.